\documentclass[sigconf,10pt]{acmart}

\usepackage{times}
\usepackage{soul}
\usepackage{url}
\usepackage[utf8]{inputenc}
\usepackage{caption}
\usepackage{tikz}
\usepackage{amsmath,amsfonts, bm}
\usepackage{algorithm}
\usepackage{algorithmicx}
\usepackage{algpseudocode}

\usepackage{graphicx}
\usepackage{textcomp}
\usepackage{booktabs} 
\usepackage{siunitx}
\usepackage{xspace,url}

\usepackage{color}
\usepackage{colortbl}
\usepackage{multirow}
\usepackage{multicol}
\usepackage{makecell}
\usepackage{tabularx}
\usepackage{listings}
\usepackage{graphicx}
\usepackage{caption}
\usepackage{enumitem}
\usepackage{adjustbox}
\usepackage{subcaption}

\usepackage[colorinlistoftodos]{todonotes}

\setcopyright{acmcopyright}
\renewcommand\footnotetextcopyrightpermission[1]{}
\acmYear{2024}\copyrightyear{2024}
\setcopyright{rightsretained}
\acmConference{Preprint}
\acmPrice{}

\definecolor{airforceblue}{rgb}{0.36, 0.54, 0.66}
\newcommand{\net}{Threshold-Net\xspace} 
\newcommand{\neurons}{Threshold Neurons\xspace} 
\newcommand{\neuron}{Threshold Neuron\xspace} 

\newcommand{\eg}{\textit{e}.\textit{g}.~}

\author{Zihao Zheng$^{1,2,*}$, Yuanchun Li$^{1,\dagger}$, Jiayu Chen$^{1,2,*}$, Peng Zhou$^{3}$, Xiang Chen$^{2}$, Yunxin Liu$^{1}$}
\thanks{$\dagger$ Corresponding author: Yuanchun Li (liyuanchun@air.tsinghua.edu.cn).}
\thanks{* This work was done while Zihao Zheng and Jiayu Chen were interning at the Institute for AI Industry Research (AIR), Tsinghua University.}
\affiliation{$^1$ Institute for AI Industry Research (AIR), Tsinghua University  \:$^2$ School of Computer Science, Peking University  \:$^3$ LuxiTech}

\begin{document}

\title{Threshold Neuron: A Brain-inspired Artificial Neuron for Efficient On-device Inference}




\begin{abstract}
Enhancing the computational efficiency of on-device Deep Neural Networks (DNNs) remains a significant challenge in mobile and edge computing. 
As we aim to execute increasingly complex tasks with constrained computational resources, much of the research has focused on compressing neural network structures and optimizing systems. 
Although many studies have focused on compressing neural network structures and parameters or optimizing underlying systems, there has been limited attention on optimizing the fundamental building blocks of neural networks -- the neurons.
In this study, we deliberate on a simple but important research question: Can we design artificial neurons that offer greater efficiency than the traditional neuron paradigm? 
Inspired by the threshold mechanisms and the excitation-inhibition balance observed in biological neurons, we propose a novel artificial neuron model, \neurons. 
Using \neurons, we can construct neural networks similar to those with traditional artificial neurons, while significantly reducing hardware implementation complexity. 
Our extensive experiments validate the effectiveness of neural networks utilizing \neurons, achieving substantial power savings of $7.51\times$ to $8.19\times$ and area savings of $3.89\times$ to $4.33\times$ at the kernel level, with minimal loss in precision. Furthermore, FPGA-based implementations of these networks demonstrate $2.52\times$ power savings and $1.75\times$ speed enhancements at the system level. The source code will be made available upon publication.


\end{abstract}

\maketitle

\section{Introduction}
\label{section:Introduction}

Artificial intelligence (AI) has showcased exceptional performance and significant influence across diverse domains. Deep Neural Networks (DNNs) have emerged as the dominant technology within AI circles, finding widespread application in a multitude of real-world scenarios \cite{VGG, ResNet}
Furthermore, the rise of diffusion models like DALL$\cdot$E3 \cite{DALLE3} and Imagen \cite{Imagen} has unveiled remarkable content generation capabilities, sparking considerable interest and research efforts.

However, efficiency is a widely-concerned problem in the deployment of DNNs. DNNs necessitate significant computational power and memory resources, posing hardware deployment challenges. Pruning-based approaches \cite{mobicom-pruning, mobisys-pruning} reduce the number of neurons to limit the resource requirement. Quantization-based methods \cite{mobicom-quantization} use a low-precision format to represent parameters to reduce storage and computing costs. Furthermore, Neural Architecture Search (NAS) is evolving to explore efficient architectures for various tasks without relying on an excessive number of parameters \cite{ShiftAddNAS}. Certain mobile/edge scenarios, such as IoT tasks and near-sensor processing, require fast and energy-efficient AI computing. Improving the efficiency of DNNs facilitates the deployment of these models on mobile and edge devices, such as mobile phones, FPGAs, and Raspberry Pi, thereby reducing hardware demands \cite{cloud1, agileNN, FlexNN}.

Despite the aforementioned approaches, few efforts are focused on the artificial neurons - the most fundamental data processing units of DNNs. Each unit in DNNs that performs signal processing can be considered as a neuron \cite{nature-norm, leaning-pooling}. From this perspective, existing artificial neurons in modern DNNs include weighted-sum, normalization, rectifier, pooling, and so on. Among them, weighted-sum is the most dominating and inefficient one that is used in linear layers, convolution layers, and attention layers \cite{AdderNet, ShiftAdd-Net}.

Traditionally, standard artificial neurons have been designed with a top-down approach that prioritizes model accuracy, often neglecting considerations for hardware efficiency.
As accuracy improvements plateau in many domains, this presents an opportunity to explore bottom-up design approaches to develop fundamentally more hardware-efficient artificial neuron paradigms.

Two key limitations in existing artificial neuron designs present opportunities for optimization.
\textit{\textbf{Limitation 1: Heavily dependent on multiplication operations.}}
The area and power consumption of the circuit required to perform a multiplication operation are the highest among all common operations.
Reducing the number of multiplication operations can effectively improve efficiency, and completely removing multiplications from neural networks would be even better, which can significantly reduce chip complexity. \textit{\textbf{Limitation 2: The use of multiple types of artificial neurons in DNNs results in hybrid circuits in accelerator design.}} Traditional DNNs employ various types of artificial neurons, each necessitating a unique circuit design as well as distinct simulation and synthesis processes. This diversity significantly escalates the complexity and time involved in designing neural network accelerators. In contrast, adopting a unified neuron model, such as the Threshold Neuron, can streamline the design process. Standardizing on a single type of neuron eliminates the need for multiple circuit designs and additional components like normalization layers. This unified approach reduces overall complexity, thereby enhancing the efficiency of the design, simulation, and implementation processes.

Therefore, an ideal design of hardware-efficient artificial neurons should have two features. First, the ideal neurons should rely on hardware-efficient operations instead of heavy multiplications. Second, the ideal neurons should be unified for compact hardware implementation. Although there are a few pioneering studies on reducing multiplications in neural networks, such as AdderNet \cite{AdderNet}, ShiftAddNet\cite{ShiftAdd-Net}, BNN \cite{BNN}, XNOR-Net \cite{XNOR-Net}, and Matmul-free\cite{matmulfree}, they still rely on multiplications (\eg normalization) or need multiple types of neurons to be effective.

To achieve this goal, we draw inspiration from the human brain and attempt to design a unified multiplication-free neuron, since the brain is probably the most powerful and efficient processor. 
Specifically, we notice two important characteristics of the brain that might be useful for efficiency.
\textit{\textbf{Characteristic 1: Biological neurons have a threshold mechanism. }}
When the signal from the previous neuron is transmitted, if it is below the threshold, the membrane potential of that neuron will not change \cite{threshold-mechanism-1}. 
This mechanism not only effectively reduces computational power consumption, but also improves the exclusion of invalid information. \textit{\textbf{Characteristic 2: Human brains have an excitation-inhibition balance.}}
The synapses of each neuron in human brains can release different neurotransmitters, which determine whether the next neuron is excited or inhibited.
Existing works have already proved that excitation-inhibition balance is the basis of learning and execution capabilities \cite{excitation–inhibition-balance1, excitation-inhibition-balance2, excitation-inhibition-balance3}. Existing artificial neurons do not possess these two characteristics.

Based on these insights, we propose a new neuron design named \neurons. First, we abstract the threshold mechanism inside each neuron, taking the thresholds as the learnable weights. Second, \neurons use subtractions rather than multiplications to achieve interaction between weights and inputs. Third, we abstract the excitation-inhibition balance into the polarity of \neurons, enabling the modeling of neurons to reflect this balance and improving the fitting capability of \neurons. 

Additionally, \neurons are utilized to construct networks within existing architectures, resulting in the creation of the \net series.
To enhance the training convergence of \net, we use a random signal initialization mechanism. 
Because of multiplication-free designs and the threshold mechanism, the aggregation effect of each layer's output in \net is not salient. As a result, normalization is unnecessary. 
Due to the inherent non-linearity in neurons, \net does not need rectifiers (activation functions). We also use convolution sampling to replace pooling. These designs make \net extremely simple for hardware support while remaining effective in modeling.

According to our experiments on various popular vision and sensing tasks, \net can achieve State-of-The-Art (SOTA) accuracy on most tasks with much less heavy computation. Our hardware simulation and FPGA prototype demonstrate the end-to-end power efficiency of our design. We anticipate that dedicated lower-level hardware support will further harness the simplicity and efficiency of \neurons.

Overall, our contributions are three-fold: 
\begin{itemize}
    \item We propose a new type of neuron called \neurons. \neurons are multiplication-free, unified, and brain-inspired, with a proper abstraction of the threshold mechanism and the excitation-inhibition balance.
    \item We use \neurons to construct networks, thus forming \net. \net does not need normalization, rectifier, and pooling, maintaining unity. It can also use existing optimization algorithms for training and updating.
    \item We evaluate \neurons and \net on various tasks. The results show that \neurons and \net achieve SOTA performance while remaining unified and multiplication-free. Our hardware simulation shows that \neurons can achieve $7.51\times \sim 8.19\times$ power savings and $3.89\times \sim 4.33\times$ area savings at the kernel level. The FPGA-based implementation shows $2.52\times$ power saving and $1.75\times$ speedup at the system level.
\end{itemize}
\section{Background and Motivation}
\label{section:background}

\begin{figure*}
    \centering
    \includegraphics[width=1\textwidth]{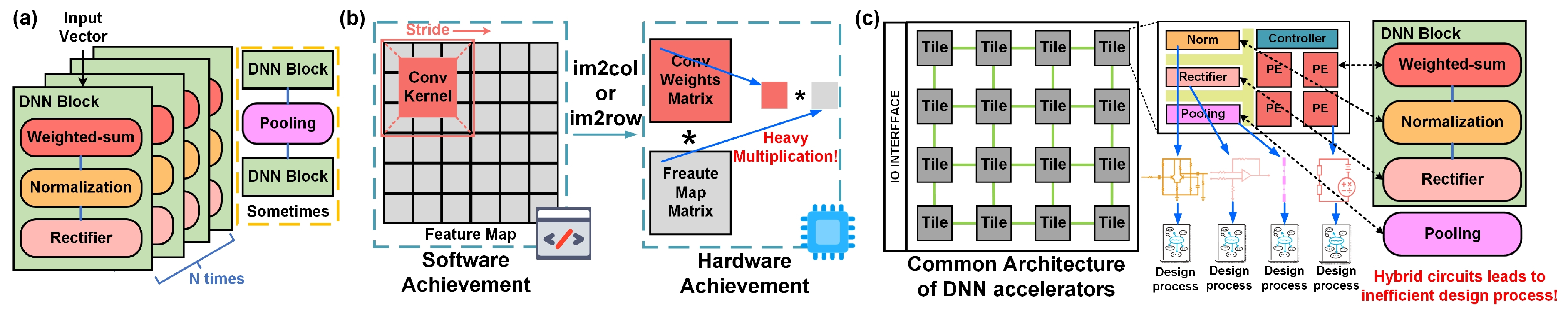}
    \vspace{-0.3cm}
    \caption{(a) The common architecture (including weighted-sum layer, normalization layer, pooling layer, and rectifier) of DNNs. 
    (b) Comparison between software and hardware implementations of convolution operations.
    (c) Hybrid circuits in DNN accelerator design.}
    \Description{figure_1}
    \label{figure_1}
\end{figure*}

\subsection{DNN Deployment on Mobile/edge Device}
Deep neural networks (DNNs) have achieved remarkable results in multiple fields such as image classification \cite{VGG, ResNet, googlenet}, object detection \cite{SSD, YOLO}, sentiment analysis \cite{DNN_for_SENTI}, and recommendation systems \cite{DNN_for_RS_1, DNN_for_RS_2}.

A DNN typically consists of multiple stacked blocks, with each block containing similar layers. For instance, as shown in Figure \ref{figure_1}(a), a typical block of DNNs consists of three parts: the weighted-sum layer (including the convolution and fully-connected layers), the normalization layer, and the rectifier layer (activation function) \cite{ResNet, googlenet}. Sometimes, the pooling layers are plugged into the block \cite{VGG}.


Weighted-sum layers are responsible for extracting features from the input data. Normalization layers, such as Batch Normalization (BN), are applied to normalize data, ensuring stable and efficient convergence of the network. Rectifiers (activation functions) introduce non-linearity, which is crucial for enabling deep neural networks to learn complex patterns. Pooling layers reduce the dimensionality of feature maps by performing down-sampling, with methods like maximum pooling selecting the highest value within a specified kernel size.





Mobile/edge devices typically have limited storage space and computing power, requiring tasks deployed on them to be low-power and efficient. DNNs are resource-intensive, so deploying DNNs on edge devices faces many problems, such as high memory overhead, slow inference speed, and high power consumption. 
Improving the efficiency of DNNs is a continuous challenge, as there is a constant demand to handle more tasks with higher quality under increasingly stringent resource constraints.

Meanwhile, the common scenarios of edge DNNs provide a different angle to rethink the optimal trade-offs in network and hardware design. For instance, many edge AI applications deployed on energy-harvesting embedded devices deal with low-dimension sensing data or low-resolution images, which does not require the high fitting ability of the most advanced AI models and operators, while the complex hardware and system for supporting the advanced models/operators may become a heavy burden in such resource-constrained devices. Therefore, an exciting research opportunity in edge AI is the integrated design of models and hardware, achieving higher whole-stack efficiency.

\begin{table}[htbp]
    \centering
    \caption{The power and area of common operations in hardware circuits. \textnormal{The frequency is limited to 50 MHz. All results come from Synopsys DC, using 28nm TSMC PDK.}}
    \vspace{-0.3cm}
    {
        \begin{tabular}{ccc}
        \toprule
        \textbf{Operations} & \textbf{Power (mW)} & \textbf{Area ($um^2$)}\\
        \midrule
        \textcolor{red}{Multiplication} & \textcolor{red}{2.48e-1} & \textcolor{red}{488.14} \\
        Addition & 1.77e-2 & 58.80 \\
        Subtraction & 1.77e-2 & 58.80 \\
        Shift & 4.8e-3 & 28.31 \\ 
        OR & 7.12e-3 & 36.06 \\
        AND & 5.93e-3 & 36.06 \\
        NOT & 5.76e-3 & 32.93 \\
        XOR & 8.72e-3 & 40.77 \\
        XNOR & 8.70e-3 & 40.77 \\
        \bottomrule
        \end{tabular}
    }
    \label{Table 1}
\end{table}

\begin{figure}[htbp]
    \centering
    \includegraphics[width=0.45\textwidth]{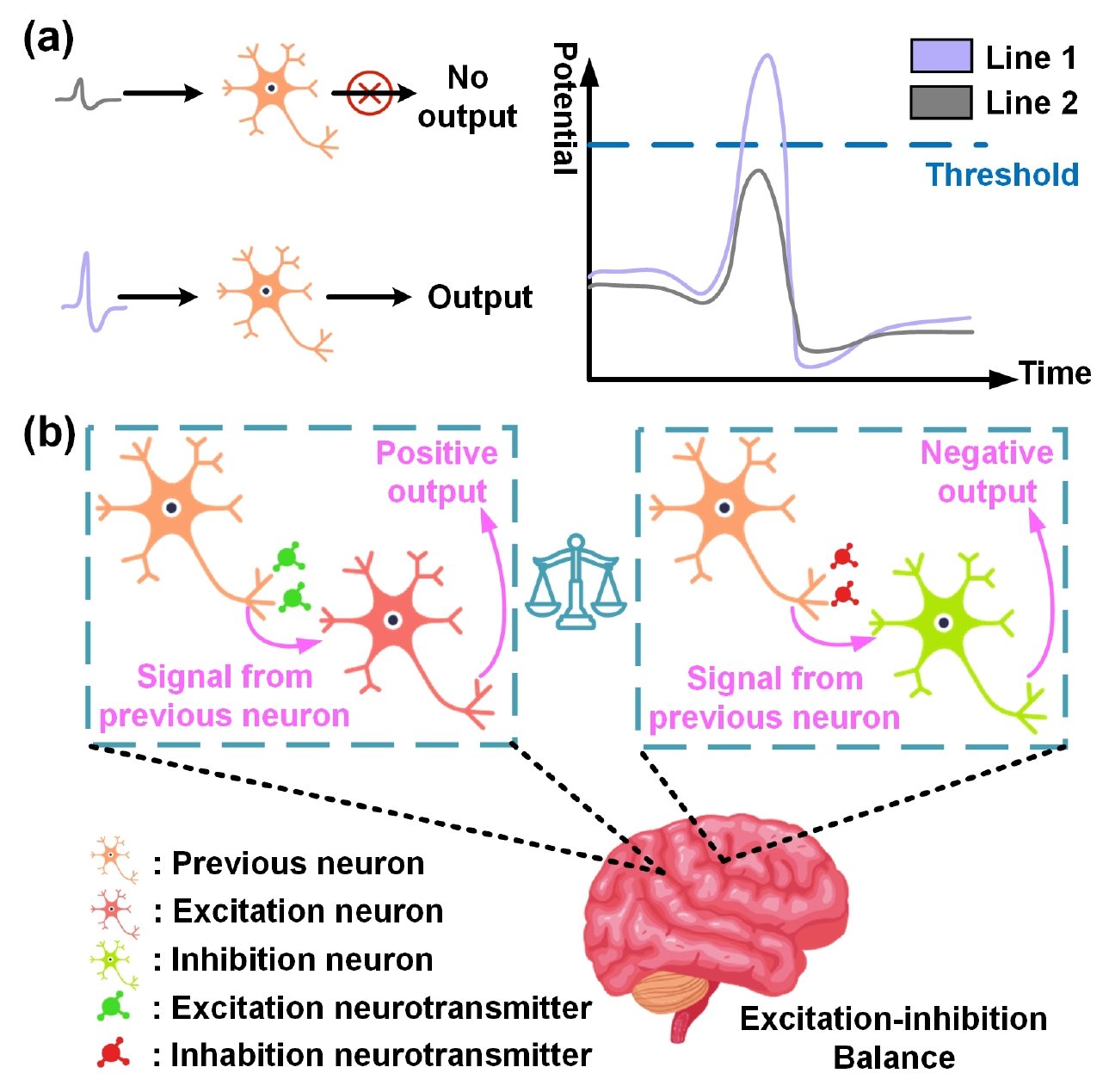}
    \vspace{-0.3cm}
    \caption{(a) The threshold mechanism in the neurons of human brains. (b) The excitation-inhibition balance in human brains.}
    \label{figure_2}
    \Description{figure_2}
\end{figure}

\subsection{Problems of Conventional Neurons}
The conventional neurons used in modern DNNs face two problems: one is the dependence on multiplications, and the other is the hybrid circuit caused by various neurons.

\subsubsection{Dependence on Multiplications}
The weighted-sum layers are the most common in DNNs, containing convolution layers and fully-connected layers. For fully-connected layers, the outputs are obtained by multiplying the weight matrix with the input matrix.

In hardware implementations, there are no specialized circuits for directly handling convolution operations with stride windows. As a result, convolution operations are typically converted into equivalent matrix multiplications using the im2col or the im2row algorithm \cite{im2col, im2col2}, as shown in Figure \ref{figure_1} (b). This conversion makes multiplication the predominant operation in DNNs.

We select typical computer operations, implement their corresponding circuits, and report their area and power consumption, as shown in Table \ref{Table 1}. 
Table \ref{Table 1} shows that the area and power consumption of multiplication circuits are significantly larger than those of other operations.
Therefore, reducing multiplications can save hardware resources and achieve efficient and green AI computing.

\subsubsection{Hybrid circuits in accelerator design}
Here, we gaze at DNNs from a bottom-up view. Each signal processing unit in DNNs can be seen as a type of neuron. 
From this perspective, common DNNs comprise multiple types of neurons, including weighted-sum, normalization, pooling, and rectifier. 
This variety leads to various hardware circuit prototypes. Each circuit prototype needs a corresponding design process, including simulation, synthesis, routing, placing, etc.

A practical instance is in SOTA DNN accelerators \cite{PRIME, ISAAC}, Processing Elements (PEs) act as weighted-sum neurons, incurring the main computing cost. 
Reusing PEs enables a large number of weighted-sum neurons using only one circuit prototype and design process. 
Nonetheless, even when the computational cost of other neuron types is low, it remains necessary to design separate modules and undertake distinct design processes for each.
hybrid circuits make the accelerator design inefficient and complex due to various types of neurons, as shown in Figure \ref{figure_1} (c). 
If the neurons inside DNNs can be unified (without other types of neurons), it will bring convenience to accelerator design.

\subsection{Human Brain Inspirations}
Human brains are powerful and efficient. Although the mechanisms leading to such efficiency are unclear yet, we notice two inspirational characteristics. One is the threshold mechanism, and the other is the excitation-inhibition balance.

\textbf{Threshold Mechanism.}
Neurons in human brains have a threshold mechanism \cite{threshold-mechanism-1}. When a neuron receives a signal input from presynaptic neurons, it compares the signal with the threshold. If the signal amplitude exceeds this threshold (like Line 1 in Figure \ref{figure_2} (a)), the neuron generates an output. Conversely, if the signal amplitude is below the threshold (like Line 2 in Figure \ref{figure_2} (a)), no output is produced. This threshold mechanism enables the neuron to effectively filter out irrelevant or low-intensity signals, thereby enhancing the efficiency of individual neurons and the overall neural network.

\textbf{Excitation-inhibition Balance.}
There are two kinds of neurons in human brains: excitation neurons and inhibition neurons.
Excitation neurons output positive signals and promote the electrical activity of other neurons through their activity. When excitation neurons transmit signals to other neurons, the output positive signal makes it easier for the receiving neuron to generate action potentials. This effect is usually achieved through the release of excitation neurotransmitters \cite{excitation-inhibition-balance2}.

Inhibition neurons output negative signals and their function is to reduce the electrical activity of other neurons. When inhibition neurons send signals to other neurons, the output negative signal makes it more difficult for the receiving neuron to generate action potentials. This effect is usually achieved through the release of inhibition neurotransmitters \cite{excitation-inhibition-balance3}.

Human brains must have both excitation neurons and inhibition neurons simultaneously and maintain their balance, as shown in Figure \ref{figure_2} (b).
This balance ensures that neural activity is neither overly excited, leading to abnormal states like epilepsy, nor excessively inhibited, resulting in functional impairment or loss of consciousness.
The balance between excitation and inhibition also contributes to efficient neural encoding and information processing, indicating its important role in efficient and economical brain functioning \cite{excitation–inhibition-balance1}.

\subsection{Design Goal and Challenges}
The aforementioned analysis motivates us to think about an interesting and important research question: 
\textbf{How can we redesign the low-level artificial neurons in DNNs to improve efficiency?} 

Addressing this question presents several key challenges.
\textbf{Challenge 1:} How to reduce the heavy dependence on multiplication operations? Conventional artificial neurons use multiplications to achieve interaction between input signals and weights. Reducing the number of multiplications in neurons will disrupt this process, making it impossible for neurons to perform weighted interactions on input signals.
\textbf{Challenge 2:} How to unify the types of neurons in a neural network? Normalization, pooling, and rectifiers are indispensable in DNNs. Removing them could result in the invalidation of neurons, potentially causing the entire network to collapse. 
\textbf{Challenge 3:} How to retain the effective learning abilities of such a new form of neural networks? The training scheme and optimization algorithm of DNNs are based on solid theoretical foundations, which are also responsible for the effective learning capability of DNNs. To ensure that newly designed neurons and networks possess strong learning abilities, they must adhere to these theories, which restrict the space of neuron design.

Fortunately, inspired by human brains, we successfully solve all these challenges and eventually design \neurons. Our design will be detailed in Section 3.
\section{Design}
\label{sec:approach}

\subsection{Overview}
We propose a new type of artificial neuron called \neurons to address the problems of existing artificial neurons used in conventional DNNs.

First, as implied by the name, we introduce a threshold mechanism in \neurons. Each signal from previous neurons will be compared with a threshold. When the amplitude of the signal is higher than the threshold, this neuron will perform calculations and output results. Otherwise, the neuron will have no output. The threshold mechanism effectively filters out insignificant signals, thus enhancing the efficiency of the neurons. Since the major functionality of multiplication operations in conventional neurons also balances the significance of signals from preceding neurons, our design actually replaces the multiplications with threshold-based comparison and subtraction operations, which are significantly more hardware-efficient.

We further develop the \neuron to get rid of other non-uniform neural operations. After the threshold operations, all signals are aggregated with summation. The aggregated output is relatively smooth, and the corresponding gradients are also relatively stable (without the gradient explosion and vanishing effects produced by weighted-sum operations). With these benefits, networks built using \neurons do not require normalization. 
In addition, the threshold mechanism introduces inherent non-linearity in neurons. Therefore, \neurons do not need separated activation units. We use thresholds as weights of neurons to make them learnable, rather than using constant thresholds, thus enhancing the fitting ability and versatility of \neurons.


Similar to the conventional artificial neurons, we can use \neurons to construct networks and design corresponding forward computation and back-propagation mechanisms to enable effective learning. The neural networks constructed with \neurons are called \net. The inherent properties of \neurons reduce the need to use rectifiers and normalization in \net, and we further use convolution to replace pooling operations in \net, leading to a fully unified design. To make \net trainable and enhance its fitting ability, we divide \neurons into two polarities and introduce a polarity balance, inspired by the excitation-inhibition balance in human brains. We propose a random polarity initialization scheme in \net. Experiment results show that this scheme is effective and \net can be trained on different tasks, achieving SOTA performance compared to other baselines.


In the remaining part of this section, we will provide a detailed introduction to the design of \neurons and \net.

\subsection{\neurons}
\subsubsection{Inspired by The Threshold Mechanism} 
\begin{figure}
    \centering
    \includegraphics[width=0.45\textwidth]{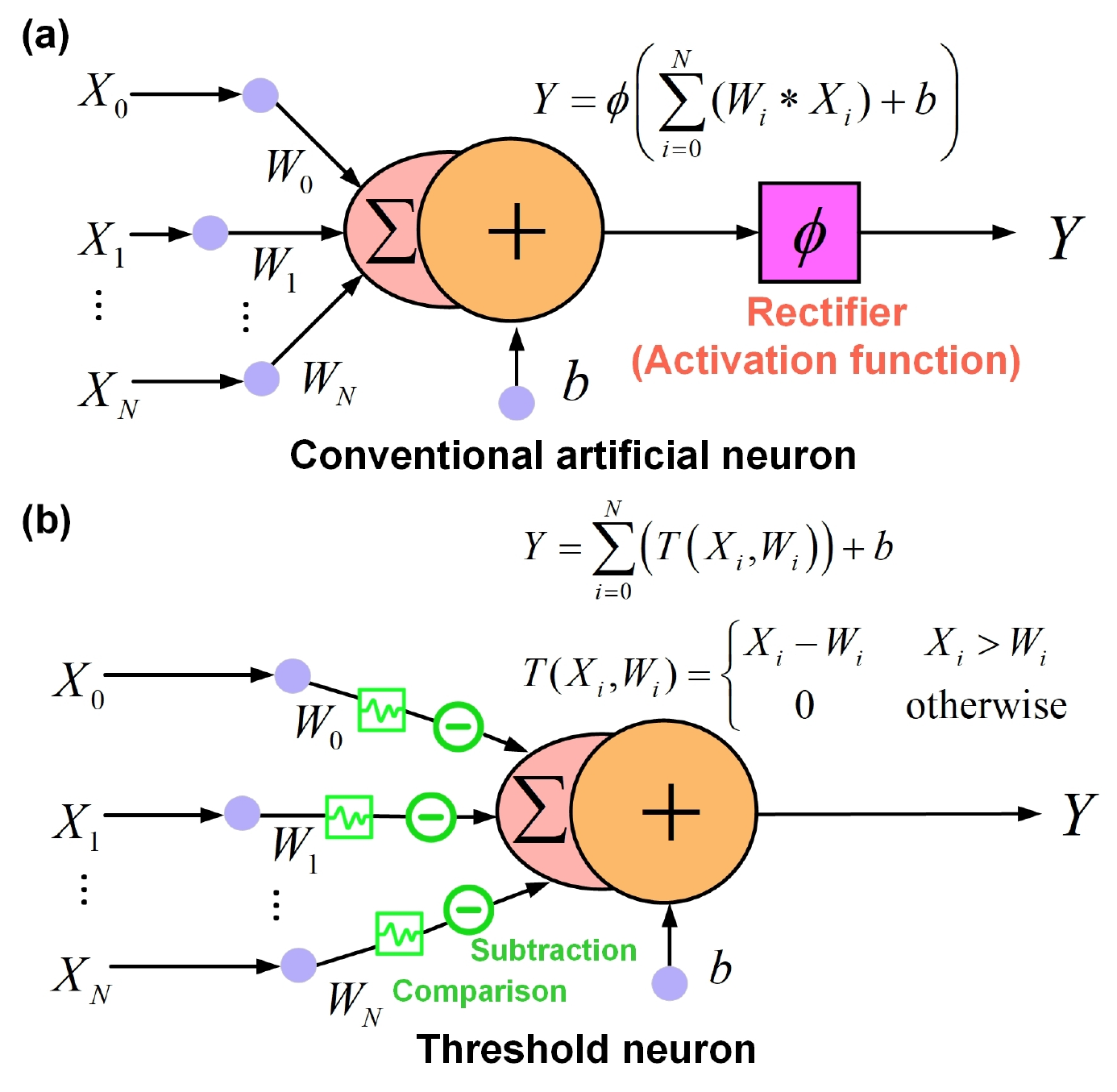}
    \vspace{-0.3cm}
    \caption{(a) The Model of Conventional Artificial Neurons. (b) The Model of \neurons.}
    \Description{figure_3}
    \label{figure_3}
\end{figure}

The model of conventional artificial neurons is shown in Figure \ref{figure_3} (a). 
This model receives signals $X_1, X_2,..., X_N$ from previous neurons. The signals multiply with corresponding weights and then are aggregated by the neuron. After adding a bias term and rectification by the rectifier, the neuron finally outputs the result. This process of artificial neurons can be represented as Formula (\ref{formula_1}). 

\begin{equation}
Y = \phi\big(\sum_{i=0}^{N}(W_{i} * X_{i}) + b\big)
\label{formula_1}
\end{equation}

Where $\phi$ represents the rectifier (activation function), $W_{i}$ represents the weight, and $X_{i}$ represents the input signal of neurons. $N$ means the number of input signals. $Y$ means the final output of the neuron, and $b$ is the bias term. 

Unlike artificial neurons, \neurons introduce a threshold mechanism and use simple operations to achieve efficient processing. The model of \neurons is shown in Figure \ref{figure_3} (b). Each input signal $X_{i}$ is compared with the corresponding threshold. We set the thresholds as the weights, making them learnable during the training process. We use subtractions to achieve the interaction between inputs and weights. The process can be written as Formula (\ref{formula_2}).

\begin{equation}
T(X_{i} , W_{i}) = 
\begin{cases}
X_{i} - W_{i} & {X_{i} > W_{i}} \\
0 & \text{otherwise}
\end{cases}
\label{formula_2}
\end{equation}

Function $T$ represents the process of comparing with the thresholds. When the amplitude of input $X_{i}$ is higher than that of $W_{i}$, the neuron will output the difference of subtraction. After $T$, signals will be aggregated, then add the bias term $b$ to get the final output $Y$, as shown in Formula (\ref{formula_3}).


\begin{equation}
Y = \sum_{i=0}^{N}\big(T(X_{i},W_{i})\big) + b 
\label{formula_3}
\end{equation}

The model and the Formula (\ref{formula_3}) show that \neurons are efficient because of using simple operations rather than multiplications. They have inherent non-linearity, eliminating the need for rectifiers. Moreover, the threshold comparisons and subtractions reduce the aggregation effect, eliminating the need for normalization.

\subsubsection{Inspired by Excitation-inhibition Balance}

\begin{figure}
    \centering
    \includegraphics[width=0.45\textwidth]{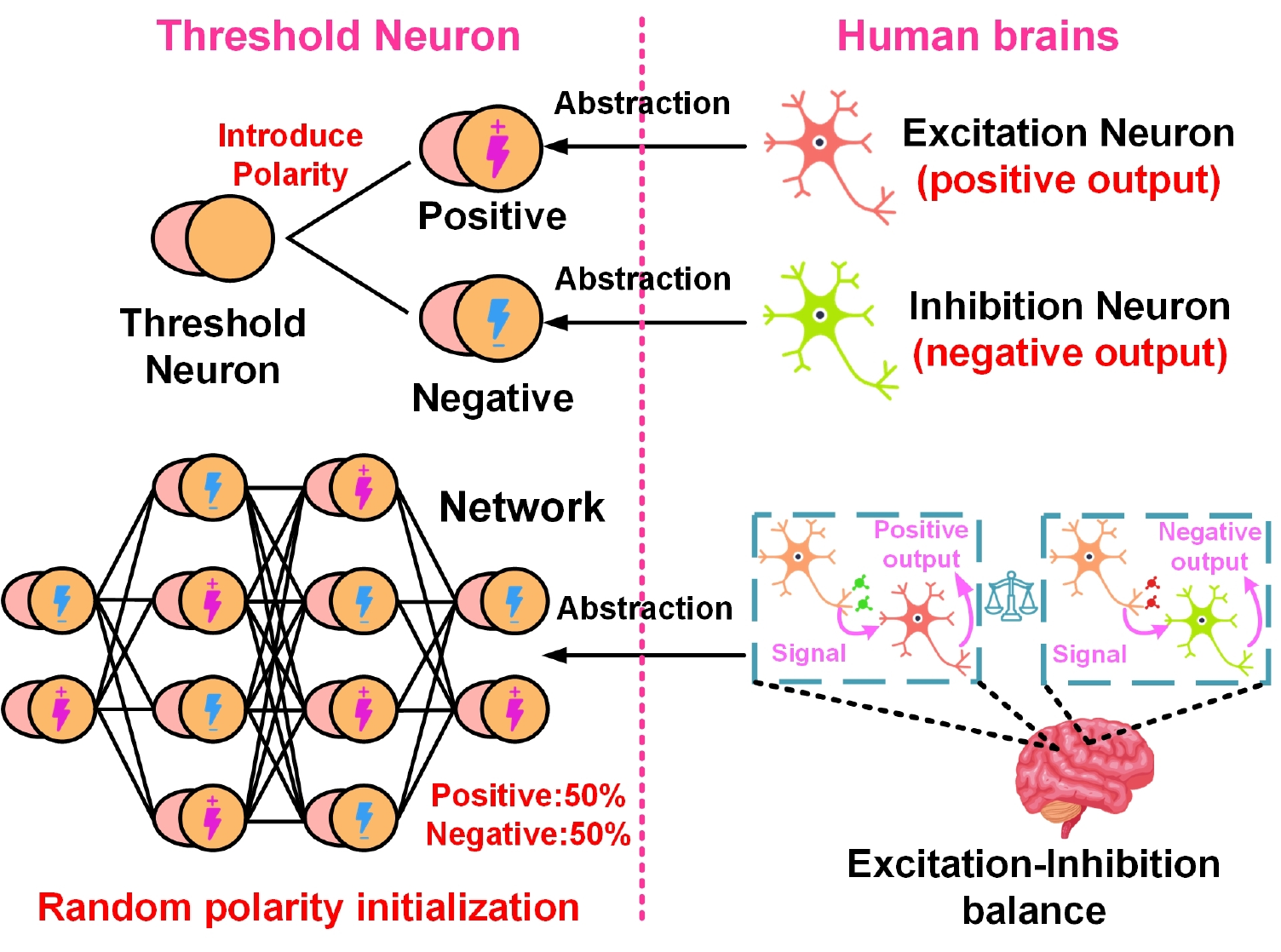}
    \vspace{-0.3cm}
    \caption{Abstraction of Excitation-inhibition Balance in \neurons.}
    \Description{figure_4}
    \label{figure_4}
\end{figure}

As aforementioned, we succeed in constructing the model of \neurons. How can we guarantee that each neuron has the appropriate fitting capability? Inspired by excitation-inhibition balance in human brains, we introduce a polarity mechanism and divide \neurons into positive and negative, as shown in Figure \ref{figure_4}. We abstract excitation neurons in human brains as positive \neurons and inhibition neurons in human brains as negative \neurons. Excitation neurons output positive signals, and inhibition neurons output negative signals. Therefore, we redefine the function $T$ of positive and negative \neurons, as shown in Formula (\ref{formula_4}) and Formula (\ref{formula_5}).

\begin{equation}
Positive: T_{pos}(X_{i}, W_{i}) = 
\begin{cases}
(X_{i} - W_{i}) & {X_{i} > W_{i}} \\
0 & \text{otherwise}
\end{cases}
\label{formula_4}
\end{equation}

\begin{equation}
Negative: T_{neg}(X_{i}, W_{i}) = 
\begin{cases}
-(X_{i} - W_{i})  & {X_{i} > W_{i}} \\
0 & \text{otherwise}
\end{cases}
\label{formula_5}
\end{equation}

In our initial design attempts, polarity initialization was crucial. Unbalanced initialization or uneven initialization can lead to adverse effects, some of which can reduce the fitting ability of \neurons, while others can make \neurons lose trainability. Considering these, we use a random polarity initialization scheme. We randomly initialize the polarity of each neuron and keep its polarity unchanged during the training process. When the number of neurons is large enough, the ratio of positive and negative neurons will be around 50\%. This scheme ensures that the polarity partition is not only balanced but also even. 

\subsection{\net}
Starting from the long-standing tradition of energy-saving hardware implementation, the design of \neurons is more hardware-efficient than multiplication-based artificial neurons. In this section, we use \neurons to construct networks and consider how to make the network trainable using the theory and algorithms of DNNs. 

\subsubsection{Forward Propagation}
Consider $F\in \mathbb{R}^{S \times S \times C_{in} \times C_{out}}$ is a filter, which is used as a layer of DNN, where kernel size is $S$, input channel is $C_{in}$ and out channel is $C_{out}$. The input figure is represented by $X \in \mathbb{R}^{H \times W \times C_{in}}$. $H$ and $W$ represent the height and width of the input figure map, respectively. The output feature map of $F \in \mathbb{R}^{S \times S \times C_{in} \times C_{out}}$ is represented by $Y$. The forward propagation of a conventional DNN layer can be written as Formula (\ref{formula_6}).

\begin{equation}
Y(m,n,t) = \sum^{S}_{i=0} \sum^{S}_{j=0} \sum^{C_{in}}_{k=0} T \big(X(m+i, n+j, k), F(i,j,k,t)\big)
\label{formula_6}
\end{equation}

$T(\cdot, \cdot)$ is a pre-defined similarity measure.
If cross-correlation is taken as the metric of distance, \eg $T(x,y) = x \times y$ and $S\neq1$, the Formula (\ref{formula_6}) transfers to the convolution operations.
When $S=1$ and $T(x,y) = x \times y$, Formula (\ref{formula_6}) represents the fully-connected layer. $T(x,y)$ of positive \neurons and negative \neurons can be represented as Formula (\ref{formula_4}) and Formula (\ref{formula_5}).

In the design of \neurons, we introduce the polarity and divide the \neurons into positive and negative. Therefore, when using \neurons to construct \net, there are two types of neurons, leading to different $T$. We use $T_{r}$ to unifiedly represent $T_{pos}$ and $T_{neg}$. So the Formula (\ref{formula_6}) can be written as Formula (\ref{formula_7}). When constructing \net, the number of neurons is large enough, so the probability distribution of $T_{r}$ can be expressed as $P(T_{r} = T_{pos}) = 0.5$ and $P(T_{r} = T_{neg}) = 0.5$. 

\begin{equation}
Y(m,n,t) = \sum^{S}_{i=0} \sum^{S}_{j=0} \sum^{C_{in}}_{k=0} T_{r} \big ( X(m+i, n+j, k), F(i,j,k,t) \big ) 
\label{formula_7}
\end{equation}

From these, we deduce each layer's computing process of \net. Formula (\ref{formula_7}) shows that \neurons are compatible with existing forward propagation theory in layer-grained.

\subsubsection{Backward Propagation}

DNNs use backward propagation to compute the gradient of each layer, and then use optimization algorithms (\eg Stochastic Gradient Descent) to update the weights. In DNNs, the backward propagation of weighted-sum layers can be represented as:

\begin{equation}
\begin{aligned}
\frac{\partial Y(m,n,t)}{\partial F(i,j,k,t)} &= \frac{\partial \big( X(m+i, n+j, k) \times F(i,j,k,t) \big )}{\partial F(i,j,k,t)} \\
& = X(m+i, n+j, k)
\end{aligned}
\label{formula_8}
\end{equation}

where $i \in [m, m+S]$ and $ j \in [n, n+S]$. In \net, $T_{pos}$ and $T_{neg}$ are  differentiable except when $X(m+i, n+j, k) = F(i,j,k,t)$. We define the derivative of this situation as $0$ so that $T_{pos}$ and $T_{neg}$ are completely differentiable. Therefore, the backward propagation process can be represented as follows. 

\begin{equation}
\begin{aligned}
&\frac{\partial Y(m,n,t)}{\partial F(i,j,k,t)} = \frac{\partial T_{pos} \big ( X(m+i, n+j, k), F(i,j,k,t) \big )}{\partial F(i,j,k,t)} \\
& =
\begin{cases}
\frac{\partial X(m+i, n+j, k)}{\partial F(i,j,k,t)} - 1 & X(m+i, n+j, k)>F(i,j,k,t) \\
0 & \text{otherwise}
\end{cases} 
\end{aligned}
\label{formula_9}
\end{equation}

\begin{equation}
\begin{aligned}
&\frac{\partial Y(m,n,t)}{\partial F(i,j,k,t)} = \frac{\partial T_{neg} \big ( X(m+i, n+j, k), F(i,j,k,t) \big )}{\partial F(i,j,k,t)} \\
& =
\begin{cases}
1 - \frac{\partial X(m+i, n+j, k)}{\partial F(i,j,k,t)} & F(i,j,k,t)>X(m+i, n+j, k) \\
0 & \text{otherwise}
\end{cases}
\end{aligned}
\label{formula_10}
\end{equation}

Therefore, \net can seamlessly integrate with the backward propagation theory of DNNs. 
\net can be incrementally updated and optimized using standard DNN training methods. However, some other multiplication-less networks \cite{BNN, XNOR-Net} face non-differentiable problems and need to use the Straight Through Estimator (STE) to estimate gradients. 

\subsubsection{Network Architecture}
\neurons are different from conventional artificial neurons, which makes us reconsider the architecture of \net. Overall, there are three changes in \net architecture.

\textbf{Change 1: Normalization-free.}
The aggregation process in artificial neurons leads to discrete feature maps and increased numerical variances, posing challenges for the network to effectively fit this discrete data distribution. Hence, normalization of the feature maps becomes essential to facilitate the network in capturing the data distribution accurately.
In contrast, \neurons employ subtractions and thresholds before aggregation, which inherently reduce the numerical disparities. As a result, normalization is unnecessary in \net. Given that normalization does not alter the tensor dimension, we opt to eliminate normalization directly from the architecture of \net.

\textbf{Change 2: Rectifier-free.}
In DNNs, rectifiers must be added between layers to introduce non-linearity. However, \neurons have natural non-linearity, so rectifiers are not required in \net. Rectifiers are usually element-wise and do not change tensor dimension, so we remove rectifiers between layers in \net.

\textbf{Change 3: Pooling-free.}
DNNs sometimes use pooling for downsampling. There are already works trying to replace pooling in DNN architecture \cite{leaning-pooling}. Inspired by this, we replace pooling layers with convolution layers in \net. To maintain the output size of feature maps, we set the convolution kernel size to equal the pooling kernel size. For other parts, we use the existing architecture of DNNs in \net.

\subsubsection{Integration with Conventional Neural Layers}
Although neural networks built exclusively with \neurons can already attain a good fitting capacity with optimal efficiency, there are situations where sacrificing some efficiency for improved accuracy may be preferable. In this section, we introduce a method to achieve such adaptable trade-offs by integrating \net layers with traditional neural layers.

Given the theoretical compatibility between \net and traditional DNNs, it is feasible to blend them by mixing \net layers with conventional neural layers and training them using the same optimization algorithms. To facilitate this integration, we introduce a layer-based Multiplication Injection (MI) mechanism. This approach allows us to interchange a \net layer with a conventional neural network layer, provided that the parameters (e.g., channels, kernel size, etc.) within the layer remain consistent.

This trade-off results in a balance: the greater the MI, the reduced proportion of \neurons, leading to decreased overall network efficiency but enhanced fitting capability.
In practical scenarios, \net can be segmented at the layer level. DNN layers can be executed on conventional computing devices (\eg CPU, GPU), while \net layers can be processed on specialized hardware. From this standpoint, the utilization of MI does not compromise the unity of \net.

Based on MI, we extend \net to diffusion models.
The diffusion noise comes from a Gaussian noise generator. 
The U-Net model accepts the noise and gradually learns the noise's distribution at each time step. 
We keep the generator normal to ensure that the noise follows a Gaussian distribution.
We construct the U-Net model by \net and use MI to improve fitting capability.
In addition, the thresholds of \neurons are learnable, so they will automatically change to fit the Gaussian distribution of the noise.
If the thresholds are static, equivalent to fixed bandwidth filters, they will disrupt the Gaussian distribution of diffusion noise.
\section{Implementation}
This section will introduce how we implement \neurons and \net. 

\begin{figure}[t]
    \centering
    \includegraphics[width=0.45\textwidth]{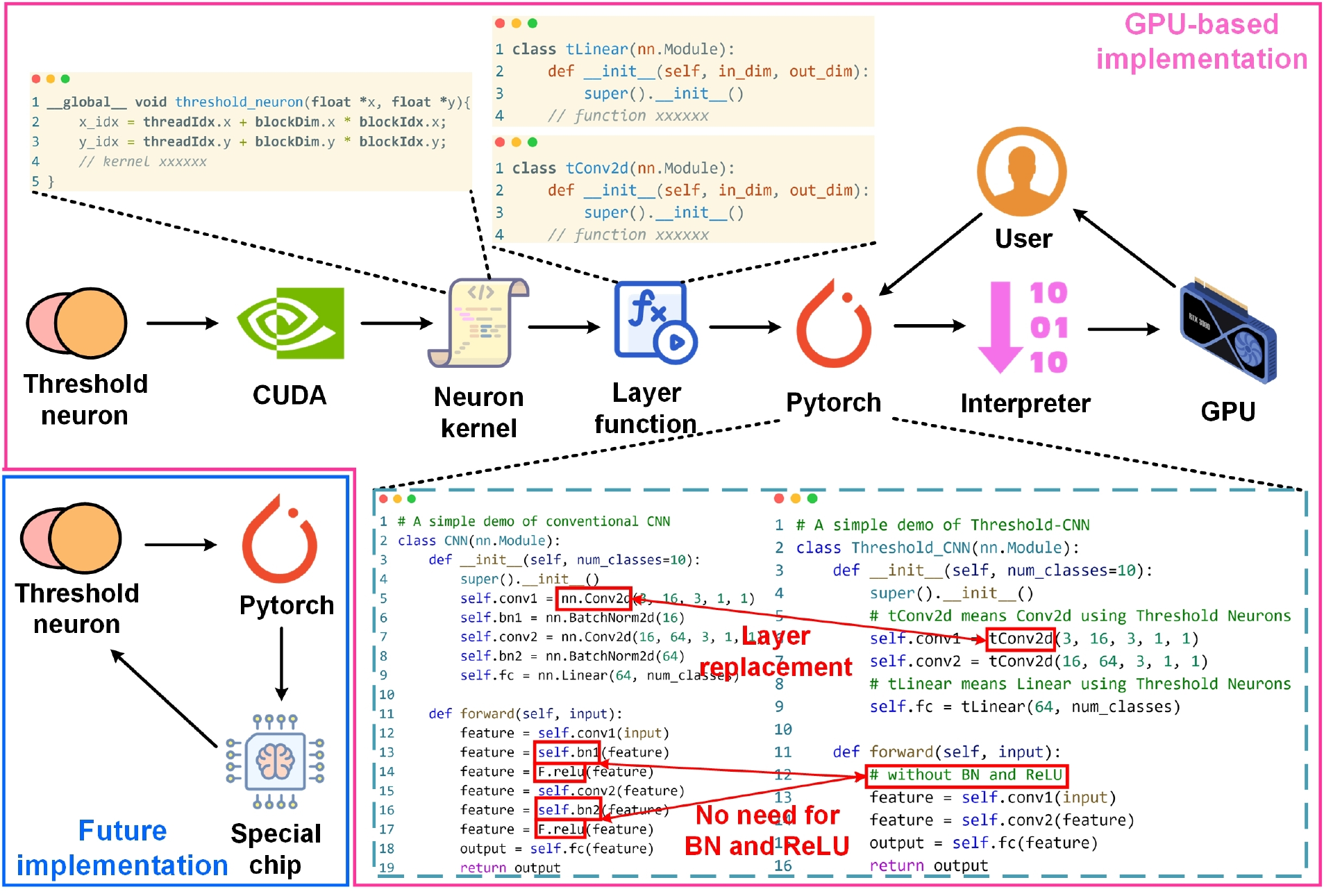}
    \vspace{-0.3cm}
    \caption{Implementation of \neurons and \net.}
    \Description{figure_implementation}
    \label{figure_implementation}
\end{figure}

\textbf{GPU Support for Training.} 
Most existing DNN models are developed and trained using the PyTorch framework on GPUs. Building on this standard, we implement Threshold Neurons as GPU kernels using CUDA to leverage the computational capabilities of modern graphics hardware. To address the complexity associated with modifying individual computing kernels in PyTorch, we implement custom layer functions analogous to the PyTorch API, namely tConv2d and tLinear. These functions facilitate the construction of our network architecture, enabling users to assemble the \net architecture in a manner consistent with traditional DNNs. Once configured, the Threshold-Net is compiled into machine code and executed on GPUs via an interpreter for training purposes. The entire process is illustrated in Figure \ref{figure_implementation}.

During the deployment phase, GPU-based implementations become suboptimal. While GPUs excel in computationally intensive tasks and matrix multiplication, \neurons do not benefit from these capabilities due to their non-reliance on multiplication operations. The ideal solution is to develop specialized hardware specifically designed for \neurons to optimize their computational paradigm. This hardware, leveraging \neurons' efficient designs, would outperform traditional GPU setups and, with robust software support similar to PyTorch, greatly enhance AI infrastructure compatibility.

\textbf{FPGA Implementation for Inference.} 
While this paper does not primarily focus on domain-specific circuit design, we contend that the inherent simplicity of \neurons offers significant advantages for hardware design. To validate this advantage, we employ an FPGA-based proof-of-concept implementation.

In GPU implementations, the CUDA kernel abstracts the functionality of weighted-sum kernels or layers; conversely, FPGA implementations are neuron-grained, operating at the level of individual neurons. Initially, we develop a single-neuron circuit using Verilog HDL.
We encapsulate the polarity of each neuron separately into a port for random initialization. Additionally, we encapsulate all thresholds in the neuron as input ports for convenient weights programming.

Due to the unity of \net, a single circuit prototype suffices for the entire FPGA implementation.
By reusing this circuit prototype, we construct various sizes of weighted-sum kernels or layers, corresponding to the instantiation in hardware design. By integrating these implemented kernels or layers, we achieve the complete implementation of \net on FPGA.

The high efficiency of FPGA implementation is manifested in two key aspects: high inference efficiency and high design efficiency.
Our experimental results, as detailed in Section 5, demonstrate a high inference efficiency.
The use of only one circuit prototype throughout underscores the implementation's high design efficiency.

\section{Evaluation}
In this section, we assess \neurons through software and hardware experiments across a range of tasks. We aim to determine if the novel neuron design can effectively address diverse modeling tasks like conventional DNNs, and investigate whether our design can enhance efficiency.
\subsection{Experimental Setup}
\subsubsection{Image Dataset}
We selected multiple image datasets to evaluate the performance of \net. 
For image classification tasks, we selected four datasets: CIFAR10 \cite{CIFAR10}, MNIST \cite{MNIST}, SVHN \cite{SVHN}, and Fashion-MNIST \cite{Fashion-MNIST}. 
For image generation tasks, we selected the MNIST \cite{MNIST} and Fashion-MNIST \cite{Fashion-MNIST} datasets as examples. 

\subsubsection{Sensing Dataset}
For near-sensor tasks, we selected six common sensing datasets: UniMiB SHAR \cite{UniMiB-SHAR}, UCI-HAR \cite{UCI-HAR}, PAMAP2 \cite{PAMAP2}, USC-HAD \cite{USC-HAD}, Daily and Sports Activities (DASA) \cite{DASA}, OPPORTUNITY \cite{OPPORTUNITY}, and WISDM \cite{WISDM}. We first segment these datasets and then pre-process the data, including data cleaning, merging, and re-formatting.


\subsubsection{Network Architecture}
We use existing network architectures to test \net, including simple CNN, ResNet18 \cite{ResNet}, and Diffusion model, shown in Table \ref{Table Net architecture}. For image classification tasks, we choose ResNet18 and modify the channel width to make ResNet18 more suitable for simple image classification tasks, reducing the risk of over-fitting. For image generation tasks, we extend \neurons to the Diffusion model. We use simple CNN for sensing tasks. For FPGA implementation, we construct a 4-layer neural network for deployment and evaluation.

\begin{table}[htbp]
    \centering
    \caption{Network Architectures Used in Evaluations of \net. \textnormal{"CONV3(64)" means kernel size is 3 and channels are 64. FC represents fully-connected layers. "BasicBlock" represents the conventional block in ResNet. "N3(5)" means a layer with 5 neurons, each neuron has 3 input channels and so on.}}
    \vspace{-0.3cm}
    \resizebox{0.45\textwidth}{!}
    {
        \begin{tabular}{ c  c }
        \toprule
        \textbf{Network} & \textbf{Architecture} \\
        \hline
        \multirow{2}*{Simple CNN} & CONV3(64)-CONV3(128)-\\
        ~ & CONV3(256)-CONV3(512)-FC-FC \\
        \hline
        4-layer NN & N3(5)-N5(3)-N3(1)-N1(1)\\
        \hline
        \multirow{2}*{Resnet18-FW64} & CONV3(16)-BasicBlock(32)\\
        ~ & -BasicBlock(64)-BasicBlock(64) \\
        \hline
        \multirow{2}*{Resnet18-FW128} & CONV3(32)-BasicBlock(64)\\
        ~ & -BasicBlock(128)-BasicBlock(128) \\
        \hline
        \multirow{2}*{Resnet18-FW256} & CONV3(64)-BasicBlock(128)\\
        ~ & -BasicBlock(256)-BasicBlock(256) \\
        \hline
        Diffusion & TimeEmbedding-UNet \\      
        \bottomrule 
        \end{tabular}
    }
    \label{Table Net architecture}
\end{table}

\subsubsection{Hardware Simulation at Kernel Level}
To demonstrate the advantages of \neurons in hardware, we construct circuits for different sizes of kernels, containing \neurons. The circuits are based on Verilog, and we use Synopsys DC for synthesis. 
We conduct power consumption and area simulations using the TSMC 28nm PDK, with a frequency limitation of 50MHz. 
For comparison, we also construct circuits for different sizes of kernels, containing conventional artificial neurons. 

\subsubsection{FPGA-based Implementation at System Level}
We use FPGA to construct a system using \neurons. 
We select Xilinx PYNQ-Z2 as the hardware platform. We use \neurons to form a 4-layer neural network and deploy its circuit prototype on FPGA. We use Vivado to measure the power consumption and latency of \neurons after place-and-route.

\subsubsection{Baseline and Evaluation Index}
For image classification tasks, we select DenseShift \cite{DenseShift}, ShiftAddNet \cite{ShiftAdd-Net}, AdderNet \cite{AdderNet}, and DeepShift \cite{DeepShift} as the baselines because they are all multiplication-less networks. We compare the accuracy, number of multiplications (Mul), and neuron types of \net with that of baselines. We also compare \net with ResNet18 under structured pruning methods to show the advantage of efficient neuron design.

For image generation tasks, we train a conventional diffusion model as a comparison and use Fréchet Inception Distance (FID) as the evaluation index. For sensing tasks, we train simple CNN and compare its accuracy with that of \net.

\subsection{Results of Image Classification Tasks}
\subsubsection{Inference Performance on GPU}
We conduct inference performance tests on Nvidia GeForce RTX 4090, with results outlined in Table \ref{Table_image_classification}. In comparison to the baselines, \net demonstrates SOTA accuracy. The baselines incorporate various neuron types, including normalization or traditional weighted-sum neurons, resulting in non-zero multiplication counts. Conversely, \net consists solely of a single neuron type. Owing to the absence of normalization, the number of multiplications in \net is reduced to zero. \neurons and \net successfully meet the design objective of efficiency and unity, contrasting with the diverse neuron types present in the baseline models.

\begin{table*}[htbp]
    \centering
    \caption{The Results of Image Classification Tasks. \textnormal{"FW" means final width of channel dimensions. Baselines use BNs or traditional weighted-sum neurons in their network architecture, thus containing multiplications. "Neuron type" means how many kinds of neurons are in the network.}}
    \vspace{-0.3cm}
    \resizebox{0.9\textwidth}{!}
    {
        \begin{tabular}{ccccccc}
        \toprule
         &\textbf{Model} & \textbf{Mul} & \textbf{Neuron type} & \textbf{Dataset} & \textbf{Acc} & ~\\
        \midrule
        \multirow{4}*{Baseline} & DenseShift-ResNet18-FW256 & 2919.44K & 5 & \multirow{4}*{CIFAR10} & \textbf{85.64\%} & ~ \\
        ~ & ShiftAddNet-ResNet18-FW256 & 118.80K & 5 & ~ & 82.45\% & ~\\
        ~ & AdderNet-ResNet18-FW256 & 2531.36K & 6 & ~ & 80.03\% & ~\\
        ~ & DeepShift-ResNet18-FW256 & 148.30K & 4 & ~ & 64.91\% & ~\\
        \hline
        \multirow{3}*{Ours}&Threshold-ResNet18-FW64 & 0 & 1 & \multirow{3}*{CIFAR10} & \textbf{83.56\%} & \textcolor{red}{$\downarrow 2.08\%$}\\
        ~ &Threshold-ResNet18-FW128 & 0 & 1 & ~ & \textbf{86.71\%} & \textcolor{green}{$\uparrow 1.07\%$}\\
        ~ &Threshold-ResNet18-FW256 & 0 & 1 & ~ & \textbf{88.53\%} & \textcolor{green}{$\uparrow 2.89\%$}\\
        \midrule
        \multirow{4}*{Baseline} & DenseShift-ResNet18-FW256 & 2230.64K & 5 & \multirow{4}*{MNIST} & \textbf{99.15\%} & ~\\
        ~ & ShiftAddNet-ResNet18-FW256 & 99.60K & 5 & ~ & 99.10\% & ~\\
        ~ & AdderNet-ResNet18-FW256 & 1942.16K & 6 & ~ & 99.04\% & ~\\
        ~ & DeepShift-ResNet18-FW256 & 98.84K & 4 & ~ & 97.82\% & ~\\
        \hline
        \multirow{3}*{Ours} & Threshold-ResNet18-FW64 & 0 & 1 & \multirow{3}*{MNIST} & \textbf{99.05\%} & \textcolor{red}{$\downarrow 0.10\%$}\\
        ~ & Threshold-ResNet18-FW128 & 0 & 1 & ~ & \textbf{99.16\%} & \textcolor{green}{$\uparrow 0.01\%$} \\
        ~ & Threshold-ResNet18-FW256 & 0 & 1 & ~ & \textbf{99.19\%} & \textcolor{green}{$\uparrow 0.04\%$} \\
        \midrule
        \multirow{4}*{Baseline} & DenseShift-ResNet18-FW256 & 2230.64K & 5 & \multirow{4}*{Fashion-MNIST} & 91.27\% \\
        ~ & ShiftAddNet-ResNet18-FW256 & 99.60K & 5 & ~ & 90.90\% & ~\\
        ~ & AdderNet-ResNet18-FW256 & 1942.16K & 6 & ~ & 90.88\% & ~\\
        ~ & DeepShift-ResNet18-FW256 & 98.84K & 4 & ~ & 87.73\% & ~\\
        \hline
        \multirow{3}*{Ours} & Threshold-ResNet18-FW64 & 0 & 1 & \multirow{3}*{Fashion-MNIST} & \textbf{91.37\%} & \textcolor{green}{$\uparrow 0.10\%$}\\
        ~ & Threshold-ResNet18-FW128 & 0 & 1 & ~ & \textbf{91.96\%} & \textcolor{green}{$\uparrow 0.69\%$}\\
        ~ & Threshold-ResNet18-FW256 & 0 & 1 & ~ & \textbf{92.13\%} & \textcolor{green}{$\uparrow 0.86\%$}\\
        \midrule
        \multirow{4}*{Baseline} & DenseShift-ResNet18-FW256 & 2919.44K & 5 & \multirow{4}*{SVHN} & 91.15\% \\
        ~ & ShiftAddNet-ResNet18-FW256 & 118.80K & 5 & ~ & 91.67\% & ~\\
        ~ & AdderNet-ResNet18-FW256 & 2531.36K & 6 & ~ & 90.52\% & ~\\
        ~ & DeepShift-ResNet18-FW256 & 148.30K & 4 & ~ & 68.03\% & ~\\
        \hline
        \multirow{3}*{Ours} & Threshold-ResNet18-FW64 & 0 & 1 & \multirow{3}*{SVHN} & \textbf{91.54\%} & \textcolor{green}{$\uparrow 0.39\%$}\\
        ~ & Threshold-ResNet18-FW128 & 0 & 1 & ~ & \textbf{92.04\%} & \textcolor{green}{$\uparrow 0.89\%$} \\
        ~ & Threshold-ResNet18-FW256 & 0 & 1 & ~ & \textbf{92.56\%} & \textcolor{green}{$\uparrow 1.41\%$} \\
        \bottomrule
        \end{tabular}
    }
    \label{Table_image_classification}
\end{table*}

\subsubsection{Training Performance on GPU}
\begin{figure}[htbp]
    \centering
    \includegraphics[width=0.45\textwidth]{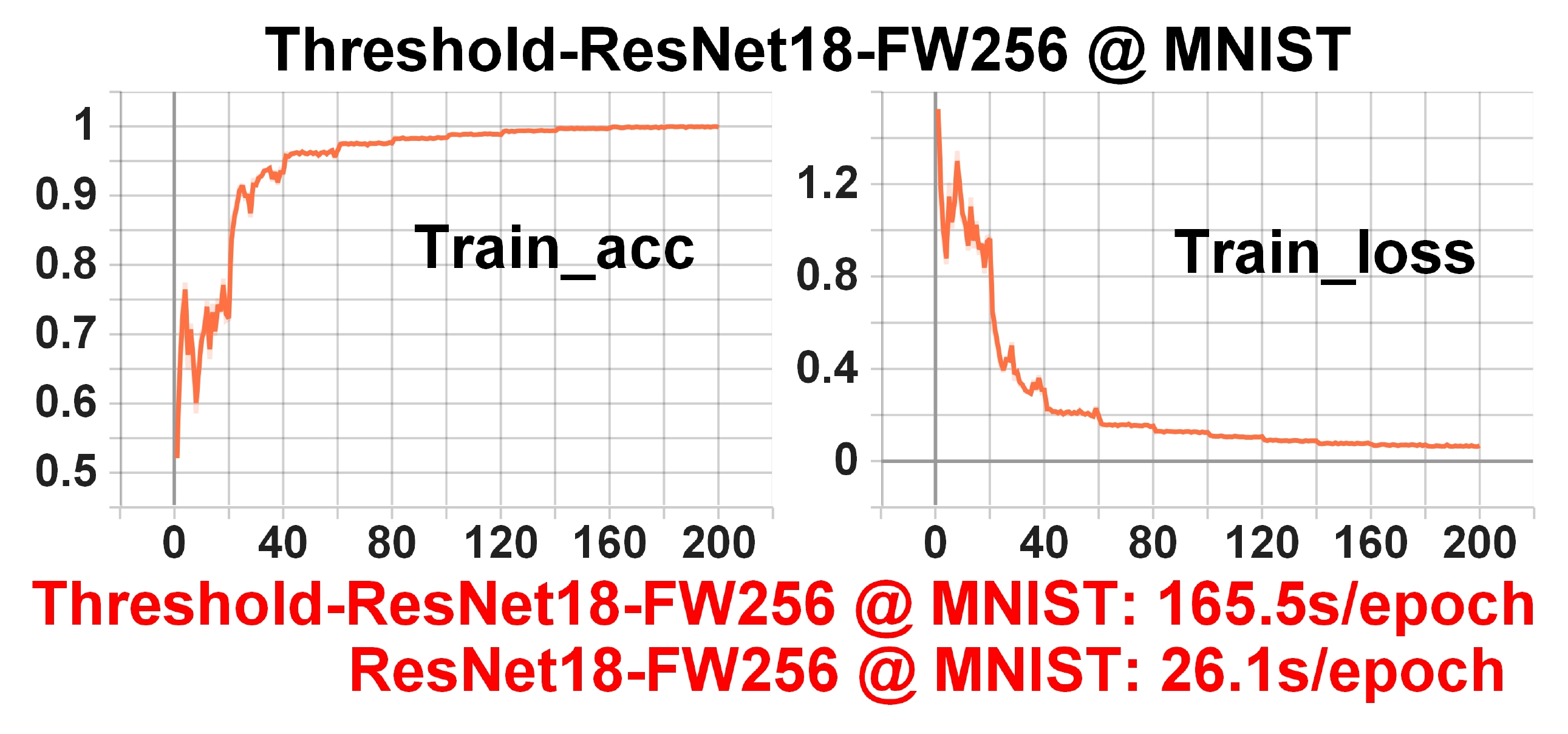}
    \vspace{-0.3cm}
    \caption{Curves of \net during Training.}
    \Description{figure_train}
    \label{figure_train}
\end{figure}

We evaluate the training performance using Nvidia GeForce RTX 4090. While we propose GPU implementation of \net, it incurs a relatively high training overhead, as illustrated in Figure \ref{figure_train}. The training speed per epoch and the convergence rate are slower than those of conventional DNNs due to inadequate GPU support. Nevertheless, given that mobile and edge devices are typically utilized for inference rather than training, our emphasis should focus on inference efficiency.

\subsubsection{Quantization}
\begin{figure}[htbp]
    \centering
    \includegraphics[width=0.45\textwidth]{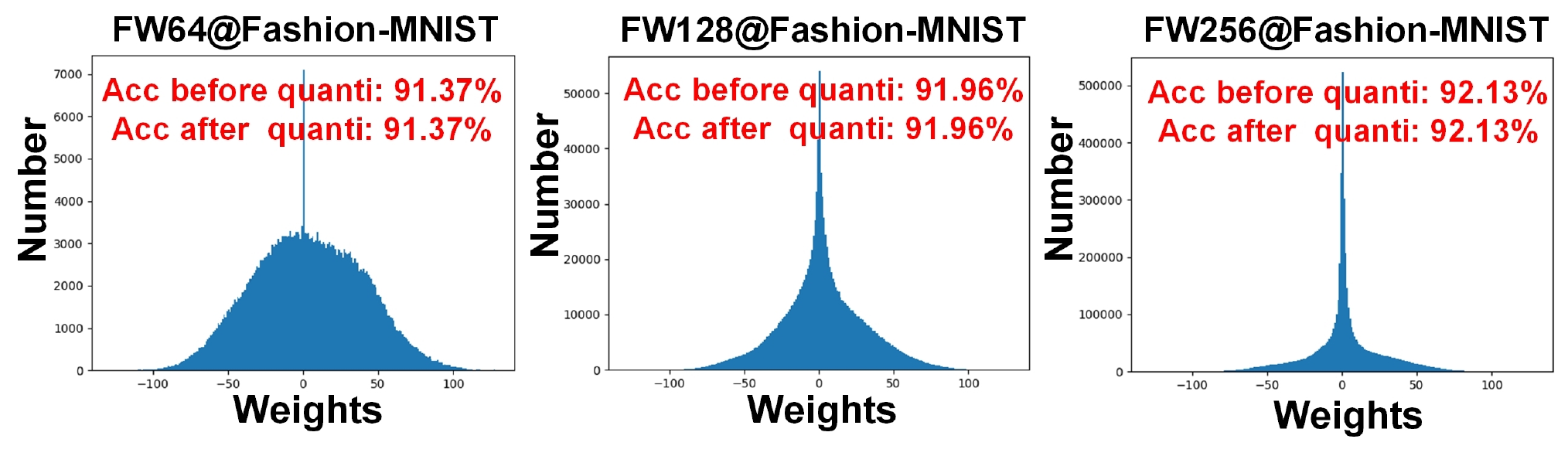}
    \vspace{-0.3cm}
    \caption{\net Weights Distribution after Quantization.}
    \Description{figure_quantization}
    \label{figure_quantization}
\end{figure}

We test the quantization of \neurons to ensure that \net can be deployed on existing hardware. We use an 8-bit uniform and symmetrical quantization method and quantize weights per tensor. The weights of artificial neurons will show as a Gaussian distribution, while the weights of subtraction-based neurons will show as a Laplacian distribution \cite{ShiftAddNAS}. From Figure \ref{figure_quantization}, the distribution of \net weights is near the Laplacian distribution. The results show that \neurons and \net are compatible with quantization methods without accuracy loss.

\subsubsection{Comparison with Pruning}
Table \ref{Table_compare_with_prune} shows the results of the comparison between \net and structured pruning. Although the pruned models have high sparsity, they still retain a large amount of multiplications, which is still inefficient for hardware. \net has no multiplications and is more efficient.

\begin{table}[h]
    \centering
    \caption{The results of comparison with pruning at CIFAR10 dataset. \textnormal{The sparsity comes from structured pruning.}}
    \vspace{-0.3cm}
    \resizebox{0.45\textwidth}{!}
    {
        \begin{tabular}{ccc}
        \toprule
        \textbf{Model} & \textbf{Acc} & \textbf{Sparsity} \\
        \midrule
        ResNet18-FW64 & 91.05\% & 0\% \\
        ResNet18-FW64 & 83.12\% & 50.75\% \\
        Threshold-ResNet18-FW64 & 83.56\% & -- \\
        \hline
        ResNet18-FW128 & 92.64\% & 0\% \\
        ResNet18-FW128 & 86.68\% & 59.90\% \\
        Threshold-ResNet18-FW128 & 86.71\% & -- \\
        \hline
        ResNet18-FW256 & 93.80\% & 0\% \\
        ResNet18-FW256 & 88.70\% & 66.43\% \\
        Threshold-ResNet18-FW256 & 88.53\% & -- \\
        \bottomrule
        \end{tabular}
    }
    \label{Table_compare_with_prune}
\end{table}

\begin{figure}[htbp]
    \centering
    \includegraphics[width=0.40\textwidth]{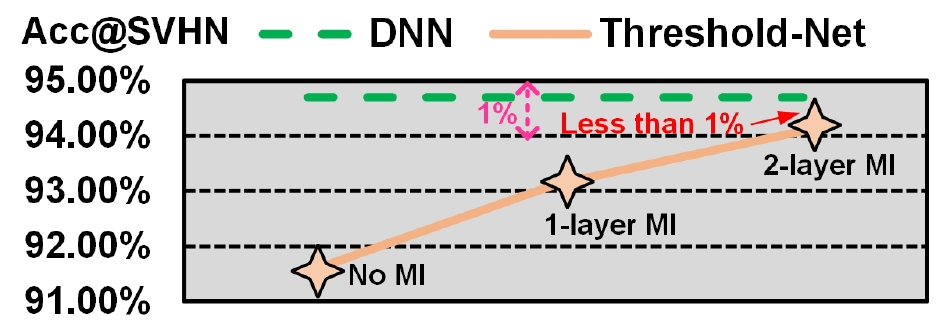}
    \vspace{-0.3cm}
    \caption{Fitting capability of \net with MI. \textnormal{"1-layer MI" means using MI to replace the first CONV3. "2-layer MI" means using MI to replace the first CONV3 and the last FC.}}
    \Description{figure_MI}
    \label{figure_MI}
\end{figure}

\begin{figure}[htbp]
    \centering
    \includegraphics[width=0.43\textwidth]{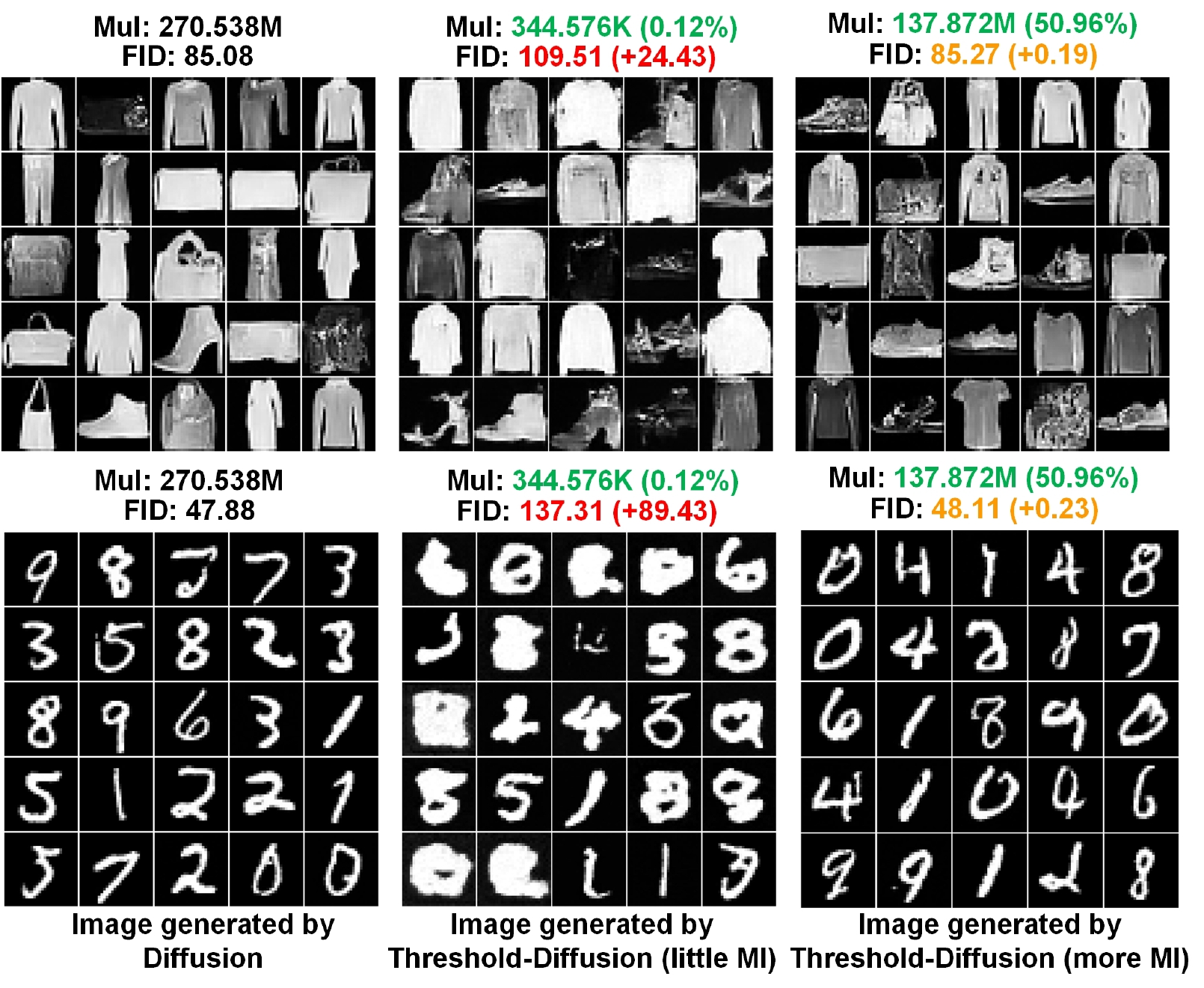}
    \vspace{-0.3cm}
    \caption{Images generated by Diffusion and Threshold-Diffusion. \textnormal{"Little MI" means using MI to replace all timestep embedding layers. "More MI" means using MI to replace both timestep embedding layers and the first layer of each block.}}
    \Description{figure_generation}
    \label{figure_generation}
\end{figure}

\subsubsection{Fitting Capability with MI} MI mixes the \neurons and multiplication-based neurons, enhancing the fitting capability of \net. We test the improvement of fitting capability with layer-wise MI, as shown in Figure \ref{figure_MI}. 
The results demonstrate that by applying MI to just two layers (the first and the last), \net can nearly match the fitting capability of multiplication-based networks, with less than a 1\% discrepancy.

\subsection{Results of Image Generation Tasks}
Due to the complexity of the generation tasks, we use MI in Threshold-Diffusion. 
When we use little MI in \net, the FID of generated images increases significantly. However, when we use more MI in \net, the FID can be recovered to its original level (Diffusion model using artificial neurons), as shown in Figure \ref{figure_generation}. That means \net with proper MI mechanism can achieve complex tasks. Moreover, \net only uses $50.96\%$ multiplications, showing its high efficiency.

\subsection{Results of Sensing Tasks}
The results of sensing tasks are shown in Table \ref{Table HAR}. In four sensing datasets, Threshold-CNN shows improved accuracy in some datasets, while in two cases, it experiences accuracy degradation of 1.37\% and 2.51\%, respectively.
In addition, Threshold-CNN is hardware-efficient and effective, which is more suitable for near-sensor computing scenarios on embedded devices.

\begin{table}[t]
    \centering
    \caption{Results of Sensing Tasks.}
    \vspace{-0.3cm}
    \resizebox{0.43\textwidth}{!}
    {
        \begin{tabular}{cccc}
        \toprule
        \textbf{Dataset} & \textbf{Network} &\textbf{Acc} & ~\\
        \midrule
        \multirow{2}*{UniMiB SHAR} & CNN & 82.60\% \\
        ~ & Threshold-CNN & 81.23\% & \color{red}{$\downarrow$ 1.37\%} \\
        \hline
        \multirow{2}*{UCI-HAR} & CNN & 95.21\% \\
        ~ & Threshold-CNN & 92.70\% & \color{red}{$\downarrow$ 2.51\%}\\ 
        \hline
        \multirow{2}*{PAMAP2} & CNN & 90.67\% \\
        ~ & Threshold-CNN & 91.90\% & \color{green}{$\uparrow$ 1.23\%}\\
        \hline
        \multirow{2}*{USC-HAD} & CNN & 86.51\% \\
        ~ & Threshold-CNN & 88.65\% & \color{green}{$\uparrow$ 2.14\%}\\
        \hline
        \multirow{2}*{DASA} & CNN & 85.80 \\
        ~ & Threshold-CNN & 87.24\% & \color{green}{$\uparrow$ 1.44\%}\\
        \hline
        \multirow{2}*{OPPORTUNITY} & CNN & 82.88 \\
        ~ & Threshold-CNN & 84.60\% & \color{green}{$\uparrow$ 1.72\%}\\
        \hline
        \multirow{2}*{WISDM} & CNN & 97.46 \\
        ~ & Threshold-CNN & 97.51\% & \color{green}{$\uparrow$ 0.05\%}\\
        \bottomrule
        \end{tabular}
    }
    \label{Table HAR}
\end{table}

\begin{figure}[t]
    \centering
    \includegraphics[width=0.45\textwidth]{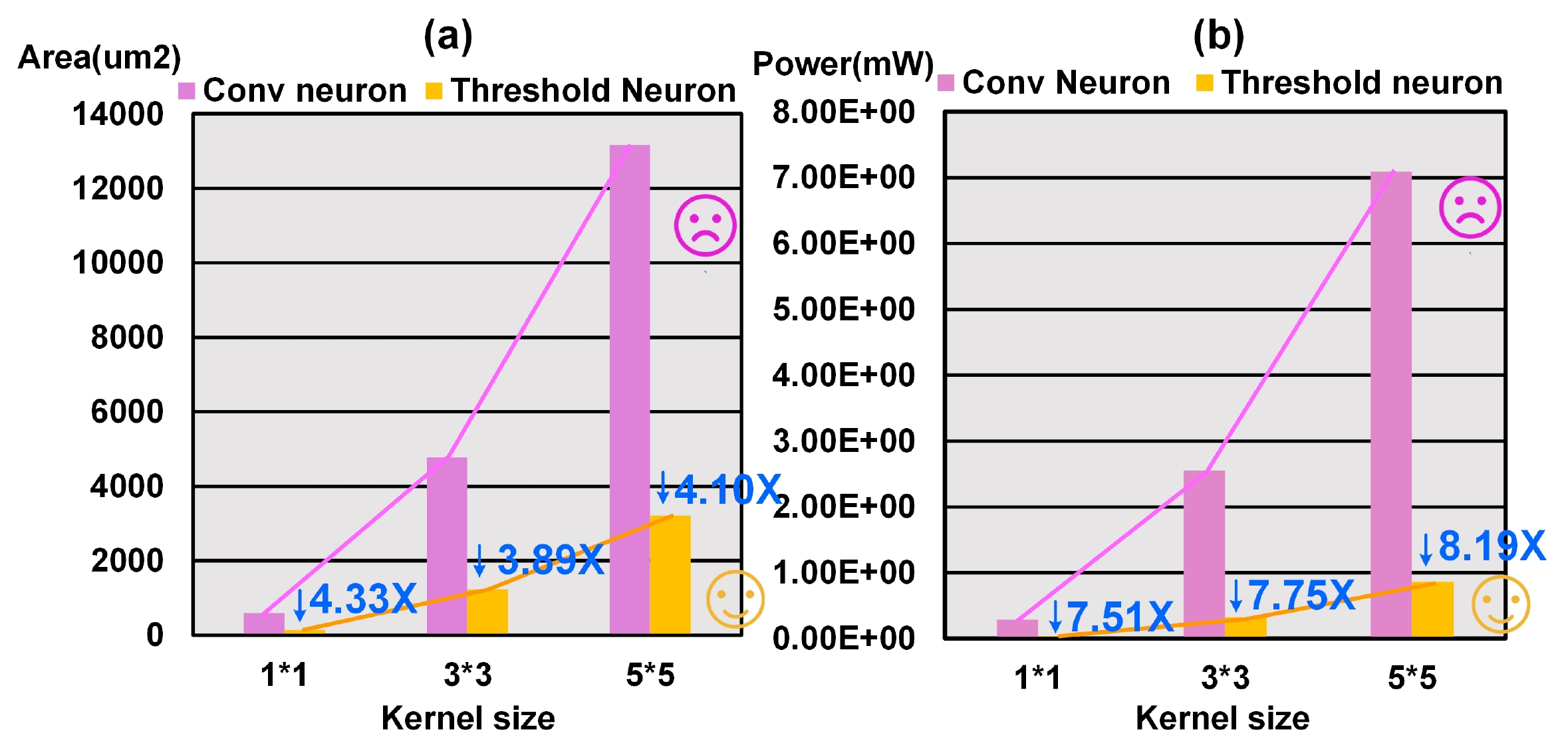}
    \vspace{-0.3cm}
    \caption{Hardware Simulation Results of \neurons. \textnormal{"5*5" means a convolution kernel whose kernel size is 5, containing 25 neurons, and so on.}}
    \Description{figure_neuron_power_and_area}
    \label{figure_neuron_power_and_area}
\end{figure}

\begin{figure}[t]
    \centering
    \includegraphics[width=0.45\textwidth]{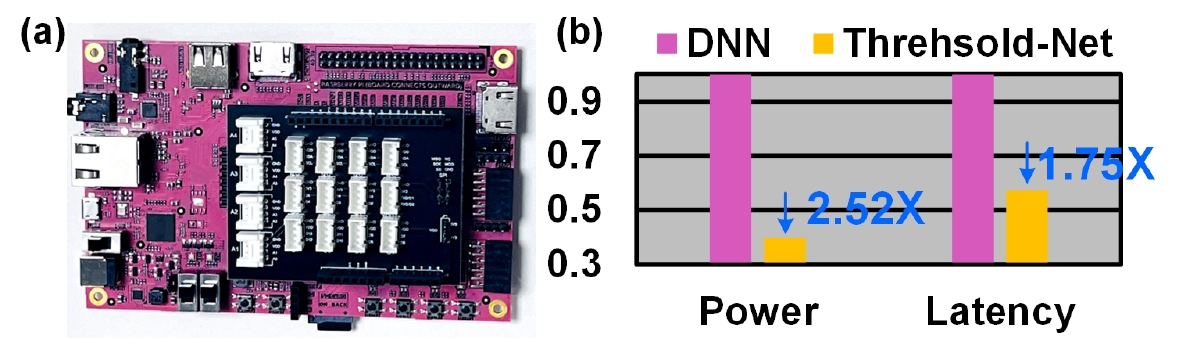}
    \vspace{-0.3cm}
    \caption{(a) PYNQ-7020 FPGA Platform. (b) Results on FPGA. \textnormal{The frequency of the system is limited to 50 MHz.}}
    \Description{figure_fpga}
    \label{figure_fpga}
\end{figure}

\subsection{Results of Hardware Simulation} 
\subsubsection{Area}
The Area results are shown in Figure \ref{figure_neuron_power_and_area} (a). We choose three kernels of different sizes and achieve their circuits. From Figure \ref{figure_neuron_power_and_area} (a), \neurons save $3.89\times \sim 4.33\times$ hardware area. Area saving allows for more space in hardware design, effectively improving the computing power of a single chip under the same area.

\subsubsection{Power}
The power results are shown in Figure \ref{figure_neuron_power_and_area} (b). \neurons save $7.51\times \sim 8.19\times$ circuit energy at kernel level. These results indicate that \neurons suit the low-power requirement of mobile/edge circumstances, leveraging less energy to achieve DNN computing.

\subsection{Threshold-Net on FPGA}
\subsubsection{Power}
Due to the multiplication-free design and the threshold mechanism in \neurons, Threshold-Net achieves $2.52\times$ power savings at the system level, as shown in Figure \ref{figure_fpga} (b). The power saving will become more salient as the number of neurons increases.

\subsubsection{Latency}
From Figure \ref{figure_fpga} (b), \net achieves $1.75\times$ speedup at the system level. As the layers get deeper, the speedup will increase. The hardware simulation is kernel-level, while the implementation of FPGA is an end-to-end system, thus leading to different results. Overall, \net shows its efficient and high-speed performance on FPGA.

\section{Related Work}
\label{sec:related_work}

\subsection{Hardware-efficient Neural Network} 
DNN models are based on heavy multiplications for feature extraction and data processing, which are high-cost and inefficient. Current works try to design hardware-efficient networks, focusing on reducing multiplications. Many works \cite{DeepShift, DenseShift, ShiftAdd-Net} use shift-based operations to replace multiplications. However, these shift-based networks require each weight to be a power of two. The weight distribution in these networks is restricted. Unlike these, \neurons use subtractions to replace multiplications rather than using shift operations.  Some works \cite{BNN, XNOR-Net} use 1-bit quantization, which can effectively simplify the multiplications. Low-bit quantization often requires retraining or other methods to estimate gradients for optimization. Unlike these, \neurons fundamentally address the basic reasons for inefficiency and are orthogonal to common quantization methods.

\subsection{On-device Inference}
With the development of AI, mobile/edge devices with sensors are gaining emerging applications, such as Human Action Recognition (HAR), and industrial fault monitoring. However, existing DNNs require huge hardware resources, posing challenges for executing inference on mobile/edge devices. Existing works \cite{FlexNN, FPGA-QHAR, agileNN} utilize model compression techniques to reduce hardware requirements. They then employ memory management or model partitioning methods to enable inference on mobile/edge devices. Yet, these works overlook the fundamental issue that multiplication-based neurons are the source of inefficiency. We propose a new type of neuron that is efficient and unified, addressing the fundamental problem.

\subsection{Brain-inspired Computing}
Brain-inspired models strive to replicate the human brain's computational paradigm to enhance computational efficiency. Spiking Neural Networks (SNNs) exemplify this approach. SNNs encode information both in the timing and rate of spikes. In contrast, \neurons are designed to handle static input-output mappings at each layer, thereby simplifying implementation while maintaining efficiency. 
In SNNs, the computation of the spike decision occurs post-aggregation of inputs. Conversely, \neurons utilize a pre-aggregation threshold mechanism, individually comparing each input to a set threshold. This method can filter out unimportant information in advance, boosting neuron efficiency. 
SNNs employ a Heaviside step function for activation, which is non-differentiable, thus requiring surrogate gradient methods \cite{surrogate} or non-gradient-based learning techniques for training \cite{STBP, STDP}. In contrast, \neurons preserve differentiability and facilitate seamless integration with standard DNNs, enabling the use of conventional optimization techniques.
\section{Discussion and future work}
\label{sec:discussion}
First, current network architectures may not be ideally suited for integrating \neurons. We utilized traditional frameworks (shown in Table \ref{Table Net architecture}) to demonstrate the potential of \net, but the optimal architecture for \neurons remains undefined. Future efforts will explore using Neural Architecture Search (NAS) and other strategies to develop more effective network configurations for \neurons.
Second, our reliance on CUDA and GPUs is a compromise since these platforms do not fully align with \neurons' operational advantages. Experiments on FPGAs reveal significant hardware efficiencies with \neurons. Developing ASICs specifically designed for \neurons could further enhance their performance, opening new pathways in hardware design tailored to their unique properties.

\section{Conclusion}
In this paper, we propose \neurons, which are hardware-efficient, brain-inspired, and unified. 
We use \neurons to construct \net within existing network architectures. Experiments have shown that \neurons and \net can significantly improve hardware efficiency while achieving commendable accuracy across diverse tasks.

\balance
\bibliographystyle{ACM-Reference-Format}
\bibliography{reference}

\end{document}